\documentclass{article}

\usepackage{arxiv}

\usepackage[utf8]{inputenc} 

\usepackage[T1]{fontenc}    
\usepackage{hyperref}       
\usepackage{url}            
\usepackage{booktabs}       
\usepackage{amsfonts}       
\usepackage{enumitem}
\usepackage{multicol}
\usepackage{graphicx}
\usepackage{mathtools}
\usepackage{float}
\usepackage{subcaption}
\usepackage[ruled,vlined]{algorithm2e}

\newcommand{\xsi}{x^{(i)}}

\DeclarePairedDelimiter{\ceil}{\lceil}{\rceil}

\title{Motion-Based Handwriting Recognition \\ and Word Reconstruction }

\author{
  Junshen Kevin Chen \thanks{All authors contributed equally to this project}\\
  Stanford University\\
  \texttt{jkc1@stanford.edu} \\
   \And
  Wanze Xie \footnotemark[1]\\
  Stanford University\\
  \texttt{wanzexie@stanford.edu} \\
    \And
  Yutong He \footnotemark[1]\\
  Stanford University\\
  \texttt{kellyyhe@stanford.edu} \\
}

\begin{document}
\maketitle

\vspace{-5px}

\begin{abstract}
In this project, we leverage a trained single-letter classifier to predict the written word from a continuously written word sequence, by designing a word reconstruction pipeline consisting of a dynamic-programming algorithm and an auto-correction model. We conduct experiments to optimize models in this pipeline, then employ domain adaptation to explore using this pipeline on unseen data distributions.
\end{abstract}

\begin{multicols*}{2}


\section{Introduction}

Handwriting recognition with vision-based approaches like Optical Character Recognition (OCR)  \cite{ocr} and on-screen stroke-detection has been successful in many applications \cite{petroski2019fully, jo2019handwritten, bib2, bib3}. However, such applications or device often require touch screens or digitizers, which are often expensive and pose an unnecessary restriction to users. Similar challenges are also identified in VR and AR environments \cite{bib4, amma2012airwriting} when user needs to recognize texts with motion sensors without a sensing screen. 

Previously \cite{ours}, we have developed a handwriting recognition system based on a simple LSTM-based model \cite{lstm} with decent performance for per-character recognition based solely on the sequential data of pen's rotation during writing. In this project, we extend individual character prediction to handwritten word reconstruction. We design a pipeline with enhanced individual character classification, dynamic-programming word candidate search, and auto-correction to tackle this problem. Moreover, we also implemented domain adaptation to explore the generalizability of our model. We find that our system work well with in-domain data, producing high accuracy of $88.8\%$ in reconstructed words. Domain adaptation boosts per-character accuracy by $43.9\%$ compared to the regular fine-tuning method.

Our code and dataset are available on Github: \url{https://github.com/deep-scribe/handwriting-recognition}.

\section{Related Work}

\paragraph{Gesture Recognition} Identifying gestures using the inertial motion unit (IMU) data has been a long standing research area in machine learning \cite{Kim_2019, ganguly2018kinect, 7418327, 6470686, 6208895}, but few studies make use of the IMU data to predict the handwriting letter due to the lack of relevant dataset. Oh et al. analyzed IMU data to recognize arabic numbers handwritten in the 3D space \cite{1363896}. However, one major problem with the system described in the study is that it requires user to wave hand in the space to outline the trajectory of the number, which contradicts with with people's habit of writing with pentip pointed down. Another work \cite{bib1} performed a similar study for recognizing handwriting digits based on pen motion. Our work shares the same spirit, but we outstand by extending the recognition to the entire English alphabet and any English words.

\paragraph{Audio Speech Recognition} We identify our work to be intrinsically similar to the Audio Speech Recognition (ASR) task, instead of to image-based tasks like OCR. This is because our goal, especially for word reconstruction, is to convert a variable data sequence into texts. Recent advance \cite{yoshimura2020end, chiu2018state, moritz2020streaming, chang2020end} in ASR has been focused on using end-to-end approaches like connectionist temporal classification (CTC) \cite{yoshimura2020end} or other sequence-to-seuqnce models like Transformer \cite{moritz2020streaming, chang2020end}. Our setting suffers from the fact we do not have any existing large scale handwriting dataset for words, and it is prohibitively expensive to collect a diverse handwriting word sequences of a large corpus from groups of people, which makes it impossible for us to explore end-to-end models. This motivates us to design a word reconstruction system based on our per-letter classification model trained with our character dataset and then test on handwriting word data.


\section{Dataset }

\paragraph{Character Sequences} 
We collect a dataset of hand-written upper-case English letters between the three of us, with each letter written 80 times to a total of 6420 sequences. A \textbf{sequence} is defined as a recording of frames of sensor values (rotation and acceleration) at a variable sampling rate of writing one letter. We collect these sequences across different recording sessions to ensure variation in handwriting positions within the individuals.

\begin{table}[H]
\centering
\begin{tabular}{lllllll}
\hline
td & yaw    & pitch  & roll   & ax     & ay      & az      \\ \hline
7  & 90.10 & -10.34 & -20.02 & 206.9 & -374.1 & 1052.9 \\
25 & 90.27  & -9.86  & -20.29 & 193.0 & -401.7 & 1046.2 \\ \hline
\end{tabular}
\vspace{3pt}
\caption{Frames from a sample of writing sequence}
\label{tab:sample-sequence}
\end{table}
\vspace{-10px}

\paragraph{Calibration and Normalization} 
To calibrate the sensor, before each recording session, the subject holds the stylus still and record for over 10 seconds, then we use the mean of the this calibration data $x^{(cali)}$ to calibrate all frames. Further, we subtract frame 0 from all frames in any sequence to get delta-rotation, invariant to stylus-holding positions. 
\[calibrated(\xsi) = \xsi - (\frac{1}{n} \sum_{j=0}^n x^{(cali)}_j) - \xsi_0\]

\paragraph{Continuous Word Sequence}
In addition to the training set of per-letter sequences, we record a set of selected 300 word sequences from 3 subjects by letting each subject write 30 specific words 3 to 4 times to produce 100 word sequences per subject. For the 30 specific words, 20 are selected from 2 English pangrams and 10 are selected from words with frequent daily usage. We use this set to test the performance of our word reconstruction system implemented based on our per-letter classifier.

\paragraph{Per-Letter Sequence Augmentation}
To overcome the issues of having a small training set, and the domain difference between individually written letters (train set) and continuously written words (test set), we apply data augmentation.

\begin{itemize}[leftmargin=*]
\item \textbf{Shape modification}: for each letter sequence, we add a Gaussian noise centered at 0 to each frame, rotate by a small quaternion vector within 5 degrees, and stretch by a random scalar $\in [1, 1.3]$ to each dimension the sequence. This generates training samples of slight variations.
\item \textbf{Prepend / append frames}: for each letter sequence, prepend a small portion of frames from the end of a random sequence, append a small portion of frames from the beginning of a random sequence, then interpolate to smoothen the transition between them. This generates samples similar to continuously written word, as the writing must move from the end of a letter to the beginning of another letter.
\item \textbf{Trimming}: for each letter sequence, randomly trim off up to 10\% of frames from its beginning and end. This generates partially written samples to simulate word reconstruction algorithm does not always slicing the word sequence perfectly.
\end{itemize}

We apply this data augmentation before \textbf{each epoch}, such that the classifier never sees the exact same sequence twice, to reduce overfitting.

\paragraph{Non-class Sequences} 
We add a 27th labeled class to the training data defined as "non-class", and ignore the sequence when the classifier predicts this label. We generate these samples by 1) taking random segments of noise from $x^{(cali)}$, 2) randomly taking sub-sequence up to $\frac{1}{3}$ of all frames of a random letter sequence. These sequences are also applied with the same data augmentation method.

\paragraph{Dataset for Auto-Correction}  
Since our auto-correct model is implemented based on SymSpell \cite{SymSpell}, we utilize the standard corpus from SymSpell for auto-correction lookup. The corpus itself is a frequency dictionary created by combining the Google Books Ngram data and Spell Checker Oriented Word Lists (SCOWL).

\section{Method}

On a high level, our pipeline splits the input word sequence into equal parts, then enumerate all possible \textbf{segments} that can make up the entire sequence. Then, we use a \textbf{character classifier} to predict the letter for each segment, producing a rank of confidence (i.e. sorting the logits in descending order), to produce a list of \textbf{candidates}. We then run a \textbf{trajectory search} algorithm to produce a list of likely \textbf{trajectories}. Finally, use the trajectories with an \textbf{autocorrect} model to produce a word prediction. We will define these terminologies in the next sub-section. The following figure demonstrates the pipeline.

\begin{figure}[H]
    \centering
    \includegraphics[scale = 0.3]{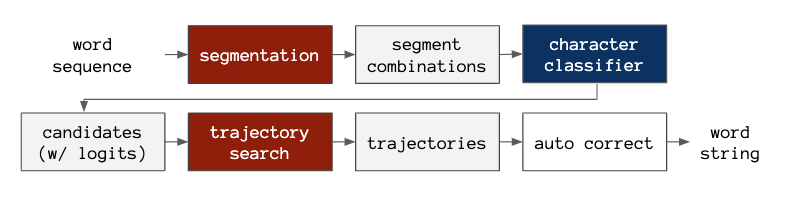}
    \caption{Word reconstruction pipeline}
\end{figure}

\subsection{Segmentation}

Intuitively, any possibly section of the word sequence can contain the writing of a letter, and thus we must first enumerate all possible segments in the word sequence to account for all possibilities. 

\begin{itemize}[leftmargin=*]
    \item \textbf{Segment}: a segment is a non-empty part of a sequence defined by a tuple of its starting and ending split-points.
    \item \textbf{Candidate}: a candidate is defined as a four-tuple \texttt{(seg-begin, seg-end, predicted-char, logit)}
    \item \textbf{Trajectory}: a trajectory is a way to slice a sequence into several candidates, each candidate corresponding to one predicted character; a trajectory must span the entire sequence.
\end{itemize}

Raw sequence that are longer should be split into more parts than shorter ones, to account for more letters possibly being written. Define granularity $G$ as number of splits on average we give to one expected letter. From our dataset, we find that the average number of frame per letter is $75$:

\begin{algorithm}[H]
\SetAlgoLined
\caption{Segmentation}
\SetKwInOut{Input}{Input}\SetKwInOut{Output}{Output} 
\Input{$W$, sequence of one written word}
\Input{$G$, granularity, num split per expected letter}
\Output{$S$, function maps segment bound to frames}
\Output{$N$, number of equal parts split}

$N \gets \ceil{len(W) / 75} * G $

$n \gets \ceil{len(W) / N} $

\For{$begin \gets 0, 1, 2, ... , N$}{
    \For{$end \gets begin , begin + 1 ,... ,N$ }{
        $S(begin, end) \gets W[begin * n : end * n]$
    }
}

yield $S,N$
\end{algorithm}

\subsection{Character classifier}
In our previous work \cite{ours}, we have experimented with different methods ranging from traditional machine learning algorithms such as K Nearest Neighbors\cite{knn}, K Means\cite{kmeans}, and deep learning frameworks such as vanilla CNN\cite{lenet} and LSTM\cite{lstm} with cross entropy loss. Among all the models we have tried, LSTM achieves the best results. 

In this project, we construct an LSTM encoder-decoder setup as the character classifier. As demonstrated in the diagram below, a multi-layer LSTM functions as an encoder to encode an input character sequence of frames. Since the output of our model is a single time-step label of 26 letters plus non-class, the decoder is a simple feed-forward structure of two fully connected layers, and the LSTM states are concatenated and flattened as input to the decoder. The output of the model is a vector of 27 logits.

To improve efficiency, we interpolate and resample the raw sequence frames into a fixed length so that we may batch the input to the model in training time and test time. The technique for resampling is base on our previous work \cite{ours}, in which we create a linear 1D interpolation model separately for each yaw, pitch, roll sequence in a writing event, and sample $N$ points along the interpolated curve for the new yaw, pitch, roll values. We use $N=100$ for our experiment.

\[\hat{Y}(n, n') \gets forward(resample(S(n, n'))) 
\atop
\forall n \in \{0,1,...,N\}, \forall n' \in \{n, n+1,...,N\}\]

\begin{figure}[H]
    \centering
    \includegraphics[scale = 0.27]{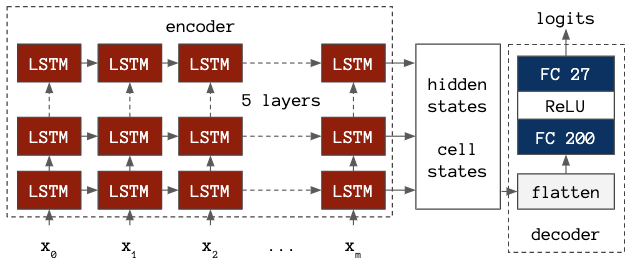}
    \caption{Character classifier architecture (example model)}
\end{figure}

\subsection{Trajectory Search}

With a per-character model prediction for each segments, we now design an algorithm that searches all combination of segments. Previously, the model produces 26 possible candidates per segment (we discard the non-class in trajectory search), and we rank and keep the top $K$ trajectories by a running average of the logits of all candidates that make up the trajectory. 

Intuitively, starting at any segment position $n$, the optimal trajectory towards the end of the trajectory is the same regardless on the trajectory arriving at $n$, and thus we leverage dynamic programming (DP) to reduce runtime complexity.

\begin{algorithm}[H]
\SetAlgoLined
\caption{TrajectorySearch}
\SetKwInOut{Input}{Input}\SetKwInOut{Output}{Output} 
\Input{$\hat{Y}$, model predictions, a function maps segment begin and end to logits}
\Input{$N$, number of equal splits in the sequence}
\Input{$K$, number of optimal trajectories to keep}
\Output{$T$, a list of $K$ optimal trajectories, sorted in descending order of average logit}

\For{$n \gets N, N-1, ...,2,1,0$}{
    \For{$n' \gets n+1, n+2 ,..., N$}{
        $\hat{y} \gets \hat{Y}(n,n')$ \\
        \For{$c \gets 0,1,2 ,...,25$}{
            $candidate \gets (n, n', c, \hat{y_c})$\\
            \For{$t \in T_{n'}$}{
                $t' \gets t \cup candidate$ \\
                $\bar{Y}_{t'} \gets \frac{\sum_{candidate' \in t'} \hat{y}_{candidate'}}{\sum_{candidate' \in t'} \mathbf{1}}$ \\
                $T_n \gets T_n \cup (\bar{Y}_{t'}, t')$
            }
        }
    }
    $Tn \gets TopK(T_n, K)$
}
yield $T_0$
\end{algorithm}

The following diagram demonstrates a contrived hypothetical example of a sequence of writing "CAT" with $N=11, K=2$. Trajectory search produces several possible trajectories ranked by their average logits. Observe that all sub-trajectories ending at position 8 may connect with all K-optimal sub-trajectories starting at position 8 (i.e. dynamic programming).

\begin{figure}[H]
    \centering
    \includegraphics[scale = 0.27]{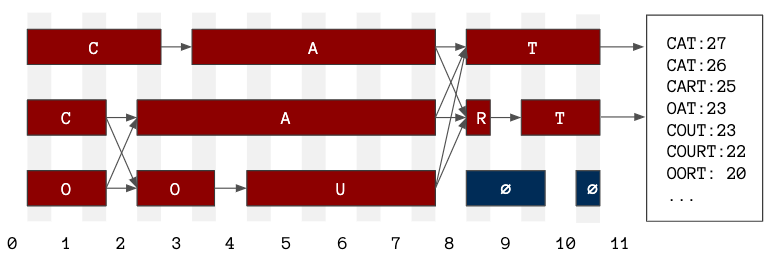}
    \caption{Trajectory search with dynamic programming}
    \label{fig:tajectory_dp}
\end{figure}

\subsection{Auto-Correction}

The last puzzle piece of our word reconstruction system pipeline is to summarize the top K predictions from the trajectory search and combine the corresponding information from the auto-correction model to finalize the prediction for the word. Our auto-correction model is based upon SymSpell, which is proven \cite{SymSpell} to be an effective non deep learning approach for correcting misspelled word. 

Contrary to directly picking the Top 1 result from the Trajectory Search and applying auto-correction, we believe that by leveraging the confidence from all Top K words as well as the information from auto-correction model, we are able to obtain a more robust reconstruction result. We experiment with four different kernel methods to find out the best way to combine the information from the Trajectory Search and auto-correction model. 

For each prediction $T_i$ from Top K results $T$, and its confidence $c_i$, we can obtain auto-correction lookup result $\hat{T}_i$, the auto-correction edit distance $d_i$, and word frequency $f_i$ depending on the corpus. Note that different $T_i$ can be the same word with different $c_i$, as shown in Figure \ref{fig:tajectory_dp}, so are the auto-correction outputs $\hat{T}_i$. Now we define four kernel functions for the auto-correction model as:

\begin{itemize}[leftmargin=*]
    \item \textbf{MaxVote}: Count the number of occurrence for each same word $\hat{T}_i$, and return the word $\hat{T}_i$ with the largest count.
    \item \textbf{SumConf}: Sum $c_i$ for each same word $\hat{T}_i$ and return $\hat{T}_i$ that has the largest confidence sum.
    \item \textbf{Division Combination}: For each auto-correction result $\hat{T}_i$, compute $\alpha_i = c_i \cdot \frac{\log{(f_i)}}{\beta \cdot d_i + 1}$, where $\beta$ is an adjustable weight to account for the magnitude difference between $f_i$ and $d_i$, and here we pick the empirical value $\beta=100$. Then we sum $\alpha_i$ for each same word $\hat{T}_i$ and return $\hat{T}_i$ that has the largest $\alpha$ sum. 
    \item \textbf{Power Combination}: For each auto-correction result $\hat{T}_i$, compute $\alpha_i = c_i \cdot \log{(f_i)}^{\frac{\beta}{d_i + 1}}$, where $\beta$ is a parameter adjusting the significance of $f_i$ which can simply be $1$, here we pick empirical value $0.75$. Then we sum $\alpha_i$ for each same word $\hat{T}_i$ and return $\hat{T}_i$ that has the largest $\alpha$ sum. 
\end{itemize}

In the experiment section we compare these four methods against simply picking $\hat{T}_0$ as the final reconstructed word. We show that leveraging the information of confidence, edit distance and frequency can significantly improve the word reconstruction accuracy.

\subsection{Domain Adaptation for Character Classification}

To compensate for the lack of generalized dataset, we designed a way to adapt feature extractor trained from the limited dataset into any new user. We seek to prove and improve the transferability and generalization ability of the model in order to make the overall pipeline more applicable to the real word, where we have limited training data, yet still would like to utilize the models for a broad range of users.


We are inspired by the domain adaptation model introduced in \cite{dann} to tackle this problem. While the LSTM encoder remains the same, we split the original decoder into two classifiers: the letter classifier and the domain classifier. The letter classifier serves the original purpose of recognizing the character written whereas the domain classifier determines whether the input sequence comes from a new subject or a seen subject. Denote sequences from the new subjects as out-of-domain (OOD) data and ones from the seen subjects as in-domain (ID) data. The feature extractor and the two classifiers are trained in a adversarial setting: the domain classifier to distinguish input data from the two domains, while the feature extractor to confuse the domain classifier while maintaining high character accuracy.

We use the original LSTM encoder and the second to the last fully-connected layer in LSTM decoder as the feature extractor, and the last fully-connected layer in LSTM decoder as the letter classifier. All pre-trained parameters are preserved at adversarial training. Then we build the domain classifier, which is a two layer fully connected network with ReLU as the intermediate activation function and sigmoid at the end. The domain classifier takes the feature embeddings extracted from the feature extractor as its input and outputs a binary prediction of ID data ($0$'s) and OOD data ($1$'s).

The loss function for training this model is defined as
\begin{align*}
    \mathcal{L}_{chr}(\theta_{chr}) &= -\sum_{i = 0}^n\sum_{c = 1}^{27}y_{c}^{(i)}\log(P(x^{(i)}|c, \theta_{chr}))\\
    \mathcal{L}_{dom}(\theta_{dom}) &= -\sum_{i = 0}^n y_{d}^{(i)}\log(P(x^{(i)}|\theta_{dom})) \\&+ (1 - y_{d}^{(i)})(1 - \log(P_{dom}(x^{(i)}|\theta_{dom})))\\
    \mathcal{L}(\theta_{chr}, \theta_{dom}) &= \mathcal{L}_{chr}(\theta_{chr}) - \lambda_p\mathcal{L}_{dom}(\theta_{dom})
\end{align*}
where $\mathcal{L}_{chr}(\theta_{chr})$ is the cross-entropy loss of the character classification and $\mathcal{L}_{dom}(\theta_{dom})$ is the binary cross-entropy loss of the domain classification. Here $y_{c}^{(i)}$ and $y_{d}^{(i)}$ are the character label and the domain label of the $i$th example respectively. $\lambda_p$ is a hyperparameter that balances the trade-off between the two objectives and $p$ represents the number of trained epoches. Here we use $\lambda_p = \frac{2}{1+exp(-10p)}-1$ to schedule the trade-off.

\begin{figure}[H]
    \centering
    \includegraphics[scale = 0.3]{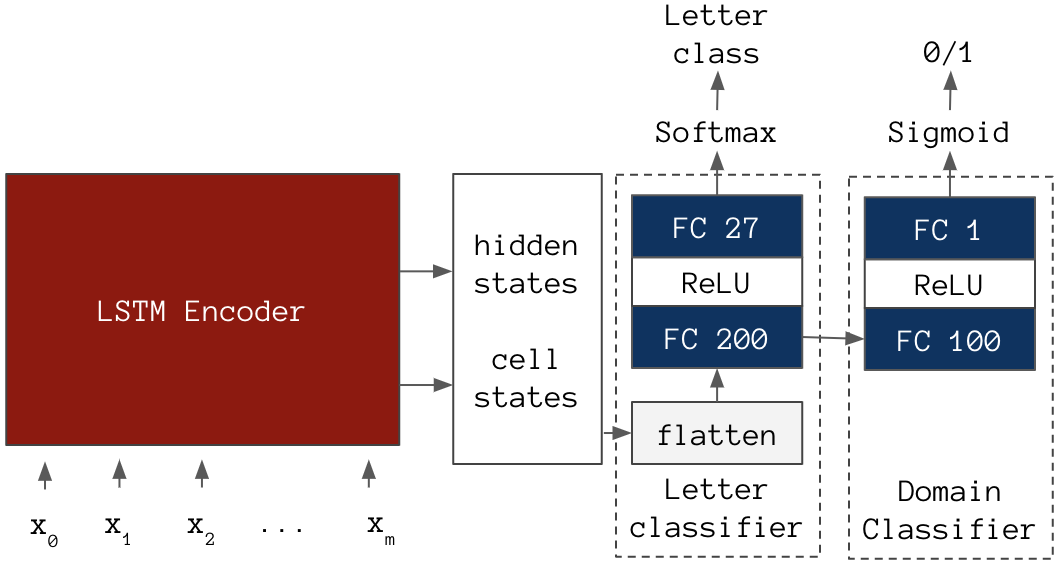}
    \caption{Domain Adaptation Architecture}
    \label{fig:da}
\end{figure}

\section{Experiments and Evaluation}

\subsection{Character classifier hyperparameter optimization}

The following hyperparameters are to be tuned for an optimal encoder-decoder architecture, we conduct random search on them within their respective integer range:

\vspace{-1mm}
\begin{itemize}[leftmargin=*, itemsep=0mm]
    \item Number of LSTM layers: $[1,8]$
    \item LSTM hidden state dimension: $[50,300]$
    \item Feed-forward hidden layer units: $[50,400]$
\end{itemize}
\vspace{-1mm}

While fixing other training parameters, such that for all models, we train with AdamW optimizer \cite{kingma2014adam} with $\lambda = 0.005$. We then rank the models by accuracy on the dev set of character sequences:

\vspace{-15px}
\begin{figure}[H]
    \centering
    \includegraphics[scale = 0.25]{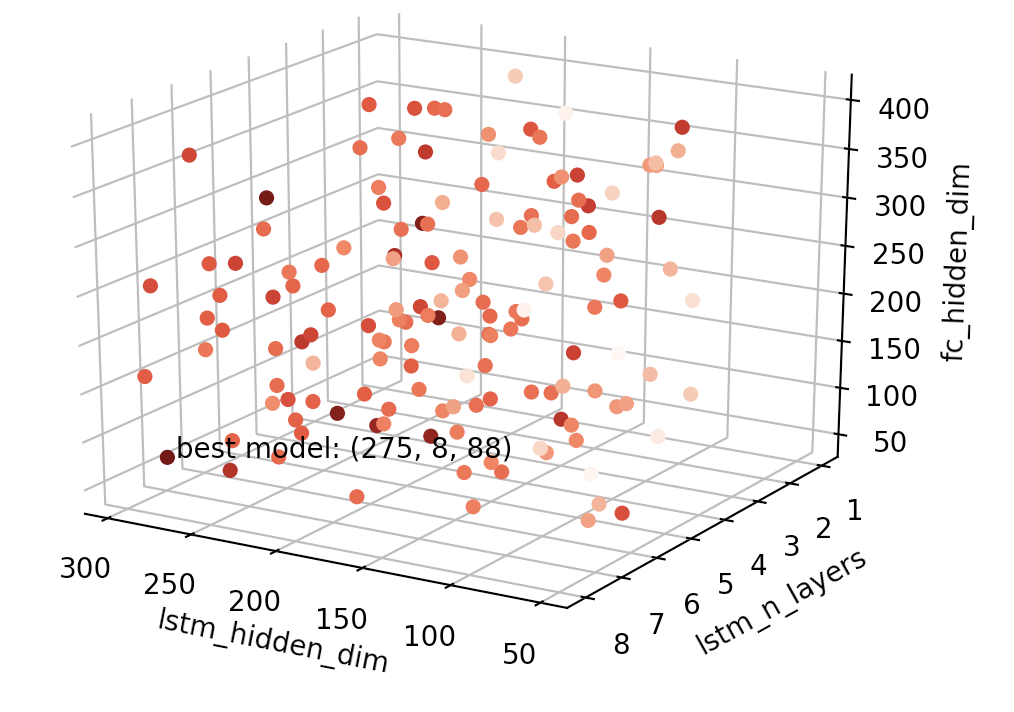}
    \caption{Dev accuracy with different hyperparameter choices. Darker color represents higher accuracy.}
\end{figure}
\vspace{-10px}

The best character classifier we find has 8 LSTM layers of 275-dimensional state, 88 feed-forward hidden units, achieving \textbf{0.9880 dev accuracy}, and \textbf{0.9931 train accuracy}.

\subsection{Word reconstruction hyperparameter optimization}

Granularity, $G$, is how many splits on average we give to one expected letter in the word sequence, at the average of 75 frames per letter. We conduct a linear search of $G \in [3,9]$. $K$ is the number of optimal "beams" we keep during trajectory search, and the number of trajectory for the auto-correct model. We conduct a search of $K \in \{5, 10, 15,20\}$. We evaluate these hyperparameters with accuracy (proportion of correct predicted word) and mean edit distance (how many letter change needed from prediction to label word).

\begin{figure}[H]
    \centering
    \includegraphics[scale=0.2]{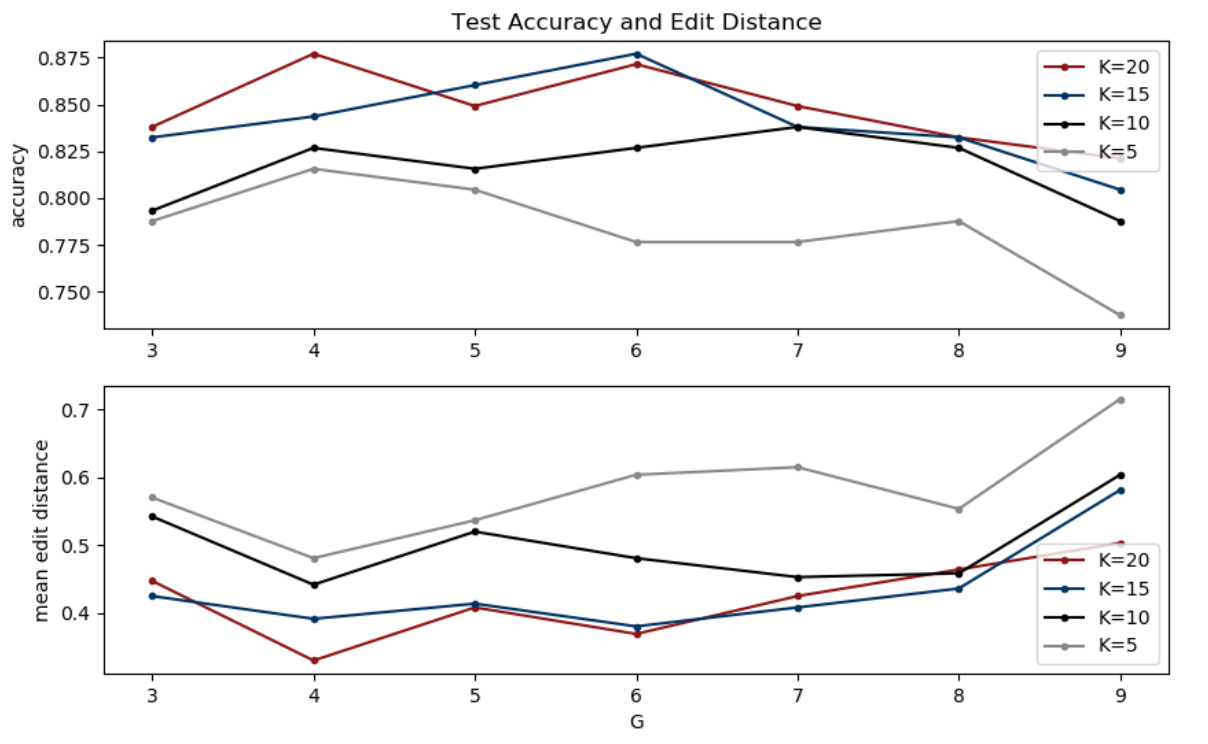}
    \caption{Metrics of hyperparameter search for $K$ and $G$}
\end{figure}
\vspace{-10px}

We find that a pipeline with $K=20, G=4$ produces the best performance of \textbf{0.8771 accuracy} and \textbf{0.3296 mean edit distance} on our test set of word sequences.

\subsection{Auto-Correction Kernel Experiment}

In order to find the best kernel methods for the auto-correction model, we conduct experiment of running the word reconstruction pipeline on the continuous word sequences testset. We use the best found model in 5.2 and choose to either directly return $\hat{T}_0$ (auto-corrected Top 1 prediction) or apply the 4 different kernel methods in the auto-correction model, creating a total of 5 experiment settings. Since our per-letter classifier is trained on the character sequences from two of the three subjects in the testset, we note the test data from these two subjects as in-domain data (Id-1 and Id-2), and the test data from the other subject as out-of-domain data (OOD). Results are shown in Table \ref{tab:ac-experiment}.  

\begin{table}[H]
\centering
\begin{tabular}{lccccc}
\specialrule{1pt}{1pt}{1pt}\\[-8pt]
  & Top 1 & MaxVote  & SumConf   & D.C. & P.C.    \\ \hline 
OOD  & 0.229 & 0.208 & 0.229 & \textbf{0.292} & 0.260  \\
Id-1 & 0.674 & 0.640 & 0.685 & \textbf{0.832} & 0.787  \\
Id-2 & 0.678 & 0.711  & 0.733 & \textbf{0.944} & 0.900  \\
Id-avg & 0.676 & 0.676  & 0.710 & \textbf{0.888} & 0.844  \\
\specialrule{1pt}{1pt}{1pt}
\end{tabular}
\vspace{3pt}
\caption{Frames from a sample of writing sequence}
\label{tab:ac-experiment}
\end{table}
\vspace{-20px}

It is observed that the kernel method Divide Combination (D.C.) performs the best across all test subjects, having 31\% improvement above the baseline with no kernel methods. We also note that while the word reconstruction system performs well on in-domain testset, it does not work for OOD subject, due to the undesirable performance of the per-letter classifier on OOD subjects, heralding the need for domain adaptation.

\subsection{Out-of-domain character classification}
To examine the effectiveness of our domain adaptation method, we split our dataset into two subsets: one ID set with two subjects' handwriting and one OOD set with one held-out subject's data. We experiment three variations of a character classifier with 3 LSTM layers of 200-dimensional state and 200 feed-forward hidden units: (1) original model trained on only the ID set; (2) original model pretrained on the ID set and then fine-tuned on the OOD set; (3) domain adaptation model pretrained on the ID set and further trained on a subset of the ID set and the entire OOD set.

The OOD dataset consists of 405 sequences and we sample 441 in-domain sequences for training the domain adaptation model. We split the datasets into training, development and testing sets with ratio 9:1:1 and train all models with AdamW optimizer with $\lambda = 0.05$ and maximum epoches $500$.

As shown in Table \ref{da_results}, domain adaptation method achieves the best results in training, development and testing accuracy and improves the other two methods with great margins.

\vspace{-10px}
\begin{table}[H]
\begin{tabular}{lccccc}
\specialrule{1pt}{1pt}{1pt}\\[-8pt]
                  & Train Acc        & Dev Acc          & Test Acc \\ \hline
Original          & \textbackslash{} & \textbackslash{} & 0.13725  \\ 
Fine-Tuning       & 0.96323          & 0.49057          & 0.49020  \\ 
Domain Adaptation &\textbf{ 0.99185}          & \textbf{0.64780}          & \textbf{0.70588}  \\
\specialrule{1pt}{1pt}{1pt}
\end{tabular}
\vspace{3pt}
\caption{Results of transfer learning and domain adaptation}
\label{da_results}
\end{table}
\vspace{-20px}

\section{Analysis and Discussion}

\paragraph{Character Classifier} From the hyper parameter search, we are able to make the preliminary conclusion that, an LSTM encoder that is deeper and higher in hidden state dimension generally performs better, while a feed-forward decoder with fewer hidden units performs better. This is likely due to the fact that the LSTM layers is required to encode a large amount of information to produce a distinctive representation across all classes, whereas for the feed-forward decoder should have fewer parameters so that it does not overfit and achieve better accuracy on unseen data. 

\paragraph{Word Reconstruction Hyperparameters}
We observe that increasing $K$ generally increases performance, and increasing $G$ increases then decreases performance within its range. A higher $K$ results in better accuracy because a large $K$ produces a bigger number of "likely correct" word trajectories ("beams") for the auto-correct model. Whereas $G$ should be tuned to a suitable value, as when it is too low, segmentation is too coarse to produce accurate slice to the word sequence between letters, and when $G$ is too high, new noise is introduced leading to false predictions.

\paragraph{Auto-Correction Kernels} 
The essential purpose of the study for Auto-Correction is to explore how we can best adapt the SymSpell algorithm in our setting, in which we have a group of candidate predictions available from Trajectory Search instead of just one. As shown in the experiment, if using the function defined for D.C. kernel to combine $c_i$ from Trajectory Search and $f_i$, $d_i$ from auto-correct lookup, we can achieve a reasonable accuracy of $88.8\%$ which enables us to deploy our word reconstruction pipeline to the real-time demo application for in-domain subjects. This demonstrates that the auto-correction module with kernel methods is an indispensable piece in the reconstruction pipeline.

\paragraph{Transfer Learning and Domain Adaptation}
As expected, the original model suffers heavily from overfitting as it only achieves $13.725\%$ on the OOD dataset. Fine-tuning the model on the OOD set improves the accuracy to $49.02\%$, which indicates that the model has the capability of learning and extracting the feature. However, since the size of the OOD dataset is designed to be small to simulate the real world application, fine-tuning is not sufficient for the model to achieve decent accuracy. With domain adaptation, the model is able to achieve $70.588\%$ accuracy, significantly improving both models. This shows that the domain adaptation model is more data-efficient and more suitable for the real world setting. However, the current result is still insufficient for trajectory search, which requires high confidence and accuracy. With an improved base model, we expect domain adaptation to help deploy the pipeline to the real word.

\section{Conclusion and Future Work}
In this project, we design a word reconstruction system that consists of a LSTM-based individual character classifier, dynamic-programming word search, and an auto-correction model. We find that the character classifier with a complex LSTM encoder and simple FC decoder works better, and trajectory search with a higher number of beams produces higher accuracy. Auto-correction with Divide Combination strategy performs the best with $88.8\%$ average word reconstruction accuracy. The domain adaptation mechanism is able to transfer the knowledge and feature extractor the model learned from the limited dataset into any new user of the device, with improvement to the base model from $13.725\%$ to $70.588\%$. 

In the future, we plan to parallelize trajectory search to speed up computation. We also plan to collect more data and further fine-tune the base character classification model to achieve better performance. With a base model trained with better data, we expect domain adaptation to be a data-efficient way to apply our system to the real world.

\clearpage

\bibliographystyle{unsrt}
\bibliography{reference}


\end{multicols*}

\end{document}